\pdfoutput=1

\documentclass[11pt]{article}

\usepackage[final]{acl}

\usepackage{times}
\usepackage{latexsym}

\usepackage[T1]{fontenc}

\usepackage[utf8]{inputenc}

\usepackage{microtype}

\usepackage{inconsolata}

\usepackage{graphicx}

\usepackage[nolist]{acronym}
\usepackage{multirow}

\begin{acronym}
    \acro{rq}[RQ]{research question}
    \acroplural{rq}[RQs]{research questions}
    \acro{nlp}[NLP]{natural language processing}
    \acro{llm}[LLM]{large language model}
    \acro{lora}[LoRA]{low-rank adaptation}
    \acroplural{llm}[LLMs]{large language models}
    \acro{ai}[AI]{artificial intelligence}
    \acro{qa}[QA]{Question answering}
    \acro{kbqa}[KBQA]{question answering over knowledge bases}
    \acro{nlg}[NLG]{Natural language generation}
    \acro{kg}[KG]{knowledge graph}
    \acroplural{kg}[KGs]{knowledge graphs}
\end{acronym}

\title{Bridging Information Gaps in Dialogues With Grounded Exchanges \\ Using Knowledge Graphs}

\author{Phillip Schneider$^1$, Nektarios Machner$^1$, Kristiina Jokinen$^2$, and Florian Matthes$^1$ \\
         $^1$Technical University of Munich, Department of Computer Science, Germany \\
         $^2$National Institute of Advanced Industrial Science and Technology, AI Research Center, Japan \\
         \texttt{\{phillip.schneider, nektarios.machner, matthes\}@tum.de} \\
         \texttt{kristiina.jokinen@aist.go.jp}}

\begin{document}
\maketitle
\begin{abstract}

Knowledge models are fundamental to dialogue systems for enabling conversational interactions, which require handling domain-specific knowledge. Ensuring effective communication in information-providing conversations entails aligning user understanding with the knowledge available to the system. However, dialogue systems often face challenges arising from semantic inconsistencies in how information is expressed in natural language compared to how it is represented within the system's internal knowledge. To address this problem, we study the potential of large language models for conversational grounding, a mechanism to bridge information gaps by establishing shared knowledge between dialogue participants. Our approach involves annotating human conversations across five knowledge domains to create a new dialogue corpus called \textit{BridgeKG}. Through a series of experiments on this dataset, we empirically evaluate the capabilities of large language models in classifying grounding acts and identifying grounded information items within a knowledge graph structure. Our findings offer insights into how these models use in-context learning for conversational grounding tasks and common prediction errors, which we illustrate with examples from challenging dialogues. We discuss how the models handle knowledge graphs as a semantic layer between unstructured dialogue utterances and structured information items.

\end{abstract}

\section{Introduction}
\label{sec:introduction}

Conversational grounding is an integral aspect of dialogues where interlocutors share information and build up a common understanding. This mutually established knowledge serves as context for subsequent interactions. For building effective dialogue systems, the \ac{nlp} community has long focused on conversational grounding, which involves inferential reasoning, dynamic feedback, and repair strategies \cite{udagawa-aizawa-2021-maintaining}. Despite extensive research, challenges remain in adapting to different conversation domains, addressing semantic vocabulary mismatches, overcoming information gaps between user knowledge and the system's internal knowledge model, as well as the lack of appropriate training data \cite{lemon2022conversational}. Owing to rapid technical advances regarding \acp{llm}, novel opportunities arise to comprehend contextual intricacies within dialogues and reconcile information expressed in natural language with that stored in machine-readable data structures.

Recognizing the limited research on \ac{llm}-based conversational grounding, we investigated the capabilities of \acp{llm} on knowledge grounding tasks. This involved annotating an existing corpus containing dialogues about different domain-specific tabular datasets. In addition to labeling grounding acts, we annotated grounded knowledge items in a knowledge graph structure, a powerful representation of complex relationships between entities and their attributes. Knowledge graphs have proven valuable in various \ac{nlp} tasks, such as disambiguating ambiguous utterances by providing contextual information \cite{hogan2021knowledge,schneider-etal-2022-decade}. For example, in dialogue systems, knowledge graphs can help identify the correct meaning of a word with multiple senses or resolve references to specific entities, enhancing the overall understanding and coherence of conversations. We opted for the JSON-LD format due to its simplicity and acceptance as a web standard, allowing interoperability by reusing existing namespaces with shared vocabularies to model knowledge from different sources and domains. 

While JSON-LD primarily uses a tree-like structure, it can represent more complex graph structures by linking nodes using identifiers like \textit{@id} and \textit{@type}. As a serialization format for Resource Description Framework (RDF) data, JSON-LD can be transformed into other formats, such as N-Triples, RDF/XML, or Turtle. This flexibility allows JSON-LD to be integrated with graph databases and other RDF tools, enhancing its utility in various applications. Table~\ref{tab:example-annotation} shows an example annotation of grounded knowledge in JSON-LD format from a conversation about nature parks.

Our contributions include (1) creating a novel dialogue corpus called \textit{BridgeKG} with over 250 conversational grounding annotations across five knowledge domains, (2) conducting a range of zero- and few-shot experiments by evaluating four \acp{llm} on two grounding tasks, and (3) summarizing common prediction errors and prompting techniques for improving model performance. To ensure the reproducibility of our experiments, we provide the \textit{BridgeKG} dataset, source code, and evaluation outputs in a public GitHub repository.\footnote{\href{https://github.com/philotron/Bridge-KG}{github.com/philotron/Bridge-KG}}

\section{Related Work}
\label{sec:relwork}
In regard to the literature on grounding in \ac{nlp}, it is essential to first define the broadly used term. Grounding can be categorized into three main types. Conversational grounding ensures a common understanding of shared knowledge within a conversation \cite{traum1994computational}. Perceptual grounding links language to sensory experiences of the real world like visual information \cite{cangelosi2010grounding}. Knowledge grounding incorporates external information sources to support \ac{nlp} systems, such as providing factual knowledge to generative language models \cite{lewis2020retrieval}.

Our study focuses solely on conversational grounding by employing \acp{llm}, a topic addressed in only a few recent studies. One related work by \citet{shaikh2023grounding} examines whether \ac{llm} generations contain grounding acts, simulating turn-taking from various conversation datasets. They found that \acp{llm} generate language with less conversational grounding than humans, often producing text that appears to assume common ground. Both their study and ours focus on the three grounding acts: explicit grounding, implicit grounding, and clarification, as proposed by \citet{clark1989contributing}. Two other closely related studies, conducted by \citet{jokinen2024towards} and \citet{mohapatra-etal-2024-conversational-grounding}, involve annotating dialogue corpora and employing language models to classify grounding acts and extract grounded knowledge items. While the former conducts preliminary experiments on two conversations with GPT-3.5-Turbo, the latter presents two annotated dialogue corpora with grounding acts, grounding units, a measure of their degree of grounding, and a baseline evaluation with the open-source T5 model \cite{raffel2020exploring}.

Unlike the mentioned related work, we are the first to conduct a series of \ac{llm} experiments aimed at knowledge identification in information-seeking conversations utilizing an in-context knowledge graph structure for identifying referenced and grounded knowledge items in dialogues.

\begin{table}[b]
\small 
\centering
\begin{tabular}{p{0.95\columnwidth}}
\textbf{Example Annotation of Grounded Knowledge}
\\
\hline
\scriptsize [\{"\textcolor{blue}{@context}": ["http://www.w3.org/ns/csvw", \{"\textcolor{blue}{schema}": "http://schema.org"\}], "\textcolor{blue}{@id}": "http://example.org/nature-parks", "\textcolor{blue}{url}": "nature-parks.csv", "\textcolor{blue}{schema:description}": "The table contains information about nature parks in Germany", "\textcolor{blue}{tableSchema}": \{"\textcolor{blue}{columns}": [\{"\textcolor{blue}{name}": "name", "\textcolor{blue}{datatype}": "string"\}, \{"\textcolor{blue}{name}": "state", "\textcolor{blue}{datatype}": "string"\}, \{"\textcolor{blue}{name}": "year", "\textcolor{blue}{datatype}": "integer"\}, \{"\textcolor{blue}{name}": "area\_in\_km2", "\textcolor{blue}{datatype}": "integer"\}, \{"\textcolor{blue}{name}": "summary", "\textcolor{blue}{datatype}": "string"\}], "\textcolor{blue}{primaryKey}": "name"\}\}, \{"\textcolor{blue}{@type}": "schema:Place", "\textcolor{blue}{name}": "Barnim", "\textcolor{blue}{state}": "Brandenburg Berlin", "\textcolor{blue}{year}": 1999, "\textcolor{blue}{area\_in\_km2"}: 749, "\textcolor{blue}{summary}": "The park includes the Barnim heath habitats dating back to the ice age. It lies between the glacial valleys of Eberswalde in the north and Berlin in the south,  and is more than half forested. The region is shaped by many individual lakes and meltwater gullies."\}]
\\
\hline
\end{tabular}
\caption{Example JSON-LD annotation of grounded knowledge from the \textit{BridgeKG} dataset, representing the system's knowledge concerning a dialogue about nature parks. Properties are displayed in \textcolor{blue}{blue} color.}
\label{tab:example-annotation}
\end{table}

\begin{table*}[h!]
\small
\centering
\begin{tabular}{l@{}ccccc@{}ccccc}
\hline
\textbf{} & & \multicolumn{4}{c}{\textbf{Zero-Shot Prompt}} & & \multicolumn{4}{c}{\textbf{Few-Shot Prompt}} \\
\cline{3-6}
\cline{8-11}
\textbf{Model} & & Accuracy & Precision & Recall & F1-Score & & Accuracy & Precision & Recall & F1-Score \\
\hline
GPT-3.5-Turbo (n=1) & & 0.64 & 0.50 & 0.46 & 0.43  & & 0.55 & 0.50 & 0.51 & 0.50 \\
GPT-3.5-Turbo (n=3) & & 0.66 & \textbf{0.81} & 0.50 & 0.50  & & 0.69 & 0.59 & 0.54 & 0.54 \\
GPT-3.5-Turbo (n=all) & & 0.59 & 0.39 & 0.44 & 0.41  & & 0.57 & 0.51 & 0.45 & 0.45 \\
GPT-4o (n=1) & & 0.39 & 0.55 & 0.54 & 0.42  & & 0.64 & 0.66 & 0.64 & 0.61 \\
GPT-4o (n=3) & & 0.59 & 0.66 & \textbf{0.67} & 0.59  & & 0.73 & \textbf{0.74} & 0.69 & \textbf{0.70} \\
GPT-4o (n=all) & & 0.64 & 0.68 & 0.66 & 0.62  & &  0.71 & 0.73 & 0.67 & 0.67 \\

Llama-3-8B (n=1) & & 0.61 & 0.54 & 0.53 & 0.54  & & 0.59 & 0.65 & 0.69 & 0.59 \\
Llama-3-8B (n=3) & & 0.65 & 0.60 & 0.60 & 0.60  & & 0.57 & 0.60 & 0.61 & 0.55 \\
Llama-3-8B (n=all) & & 0.44 & 0.55 & 0.39 & 0.38  & & 0.55 & 0.54 & 0.51 & 0.51 \\
Llama-3-70B (n=1) & & 0.41 & 0.54 & 0.56 & 0.43  & & 0.51 & 0.61 & 0.63 & 0.53 \\
Llama-3-70B (n=3) & & 0.59 & 0.66 & \textbf{0.67} & 0.59  & & 0.65 & 0.68 & 0.69 & 0.64 \\
Llama-3-70B (n=all) & & \textbf{0.71} & 0.66 & 0.64 & \textbf{0.64}  & & \textbf{0.76} & 0.70 & \textbf{0.70} & \textbf{0.70} \\
\hline
\end{tabular}
\caption{Zero-shot and few-shot performance metrics for grounding act classification evaluated by macro-averaged accuracy, precision, recall, and F1-score. The variable n denotes the number of preceding input utterances. Bold values highlight the best value for each metric.}
\label{tab:cls-results}
\end{table*}

\section{Method}
\paragraph{Dataset Annotation}
\label{sec:annotation}

The source dialogue corpus we reuse was collected in a study on exploratory information-seeking conversations from \citet{schneider2023investigating}. It comprises 26 conversations about tabular datasets on real-world knowledge spanning the domains of geography, history, media, nutrition, and sports. Every conversation involved a pair where one person was the information seeker and the other was the information provider, using a text-based chatroom for communication. The information seekers were instructed to discover and gather new information about their partner's previously unknown dataset. Two researchers annotated each written dialogue with labels for grounding acts (explicit, implicit, and clarification). Explicit grounding involves a response that clearly confirms understanding or acceptance of received information (e.g., ``okay, thanks''), whereas implicit grounding moves the conversation forward without explicitly acknowledging or questioning the recently shared information (implicit acceptance). Clarification occurs when a conversation partner seeks more information about thus far presented knowledge, which does not result in grounded knowledge since mutual acceptance has not yet been reached. 

For explicit and implicit labels, the grounded knowledge items that have been shared until this point in the dialogue were annotated as a knowledge graph structure in JSON-LD format \cite{jsonld11}. Annotation disagreements were collaboratively resolved to reach a consensus. Knowledge is incorporated into the grounding annotation only if it is a subset of the underlying tabular dataset and can be represented within the modeled internal system knowledge, which we defined using vocabulary from the namespaces \textit{Schema.org} and \textit{CSVW} \cite{csvw,schemaorg}. An example conversation illustrating labeled grounding acts and grounded knowledge items for individual dialogue utterances is provided in Table~\ref{tab:dataset-example} in Appendix~\ref{sec:appendix-a}.

\paragraph{Experimental Setup}
\label{sec:experimental-setup}
Based on the annotated dataset with conversational grounding labels, we conducted several experiments using four state-of-the-art \acp{llm}: the open-source Llama-3-8B-Instruct as well as Llama-3-70B-Instruct \cite{meta2024llama} from the Llama 3 model family, and the closed-source models GPT-3.5-Turbo (version: 0125) and GPT-4o (version: 2024-05-13) \cite{openai2022chat,openai2024gpt4o}. We defined two model prompts: one for classifying grounding acts and another for identifying grounded knowledge. For the knowledge identification prompt, which tasked the \ac{llm} to predict the grounded knowledge subset in the conversation thus far, we provided both the input dialogue and the complete system knowledge (i.e., the annotated grounded knowledge for the entire conversation). All models were prompted using a chat completion format, which included a system instruction and, in the few-shot setting, three in-context examples presented as user and assistant turns. Both model prompts are provided in the Appendix in full length (Tables~\ref{tab:prompts-cls} and \ref{tab:prompts-gk}). To promote deterministic generation, we set the generation seed to 1 and the temperature parameter to 0. The maximum token limit was set to 128 for classification and 4096 for grounded knowledge identification. All generated outputs with extra text were preprocessed using a regular expression to match and extract the first occurrence of either the grounding act or JSON-LD array.

\section{Results and Discussion}
\label{sec:results}
\paragraph{Classification of Grounding Acts}

\begin{figure}[b!]
  \includegraphics[width=\columnwidth]{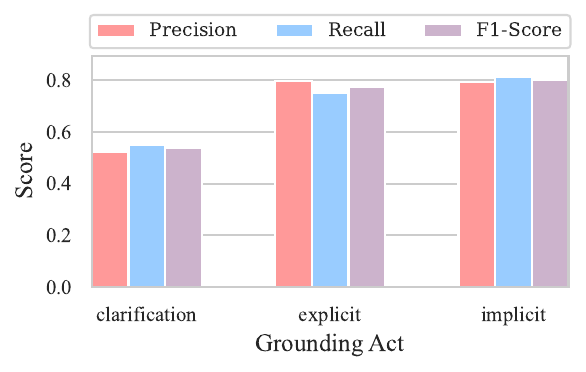}
  \caption{Performance comparison of precision, recall, and F1-score by grounding act for the Llama-3-70B model with all input utterances (n=all).}
  \label{fig:bar-grounding}
\end{figure}

\begin{table*}[t!]
\small
\centering
\begin{tabular}{lcccc}
\hline
\textbf{Issue Type} & \textbf{GPT-3.5-Turbo} & \textbf{GPT-4o} & \textbf{Llama-3-8B} & \textbf{Llama-3-70B}  \\
& \multicolumn{4}{c}{Relative Frequency: Zero-Shot / Few-Shot} \\
\hline
Invalid JSON-LD &  0.00 / 0.01 &  0.00 / 0.00 & 0.02 / 0.09 & 0.20 / 0.00 \\
\hline
Property Hallucination &  0.01 / 0.00 &  0.00 / 0.02 & 0.08 / 0.22 & 0.38 / 0.26 \\
Value Hallucination &  0.02 / 0.00 &  0.01 / 0.03 & 0.22 / 0.05 & 0.46 / 0.07 \\
\hline
Property Excess & 0.49 / 0.48 &  0.29 / 0.24 & 0.50 / 0.38 & 0.61 / 0.51 \\
Property Deficit & 0.37 / 0.22 &  0.31 / 0.09 & 0.50 / 0.36 & 0.39 / 0.20 \\
\hline
Value Excess & 0.68 / 0.63 &  0.40 / 0.31 & 0.66 / 0.32 & 0.76 / 0.47 \\
Value Deficit & 0.22 / 0.22 &  0.29 / 0.28 & 0.34 / 0.62 & 0.24 / 0.34 \\
\hline
\end{tabular}
\caption{Relative frequency of issues in zero- and few-shot predictions for grounded knowledge identification.}
\label{tab:issues-frequency}
\end{table*}

Table~\ref{tab:cls-results} shows the performance for classifying grounding acts, using macro-averages to ensure equal class importance. Nearly all tested \acp{llm} benefited from the added context of few-shot examples, with F1-scores generally improving; however, this improvement diminishes as the number of input dialogue turns (n) increases, suggesting potential redundancy when in-context examples are already provided. The results indicate that n=3 often optimizes performance in both zero- and few-shot settings by balancing context retention, noise reduction, and efficient usage of tokens. While Llama-8B's performance drops from 0.54 F1-score at n=1 to 0.38 at n=all, larger \acp{llm} like Llama-70B and GPT-4o handle longer input better, probably due to a higher parameter count and superior noise handling.

Another significant finding is the competitive performance of open-source \acp{llm} against proprietary ones: Llama-8B surpasses GPT-3.5 in the zero-shot run, and Llama-70B matches GPT-4o in the few-shot run. The breakdown of Llama-70B's performance by grounding act, illustrated in Figure~\ref{fig:bar-grounding}, reveals clarification as the most challenging act to classify, consistent with our observation of the other \acp{llm}. For instance, the models often struggled when users tried to clarify a previously introduced concept. Instead of recognizing the clarification (e.g., ``And category describes whether it is a movie, tv show, or work of literature?''), the models often misinterpreted it as introducing a new topic, falsely assuming that the previous concept is already implicitly grounded. Contrary to clarification acts, the F1-scores for explicit and implicit classification are comparable. Despite achieving the same overall F1-score, GPT-4o tends to overpredict implicit labels in contrast to the more balanced Llama-70B, as revealed by the confusion matrices in Figure~\ref{fig:heatmap-plot} in Appendix~\ref{sec:appendix-a}. The latter shows that GPT-4o excels at predicting explicit grounding accurately, avoiding false positives altogether, but it tends to overpredict the implicit class, particularly in cases where participants acknowledge information explicitly before asking a new question (e.g., ``Ok very interesting! What is the highest level of protein in the chart?'').

\paragraph{Identification of Grounded Knowledge}

The second series of experiments aimed at identifying grounded knowledge for a suitable dialogue context, which is a significantly more complex task than classifying grounding acts \cite{wu-etal-2021-dialki,oh-etal-2023-pk}. Knowledge identification required the \acp{llm} to uniquely pinpoint specific knowledge items from a set of possibilities within the system knowledge model, bridging between vague conversation utterances and structured JSON-LD arrays. 

\begin{figure}[b!]
  \includegraphics[width=\columnwidth]{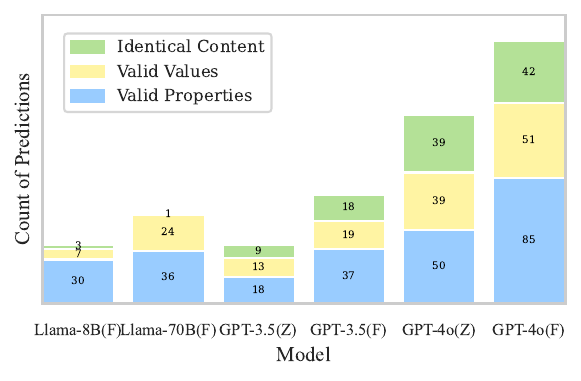}
  \caption{Count of predictions in JSON-LD format with valid properties, valid values, or identical content for evaluated models in zero- (Z) and few-shot (F) settings.}
  \label{fig:stack-grounded-knowledge}
\end{figure}

Figure~\ref{fig:stack-grounded-knowledge} depicts the count of JSON-LD generations accurately matching our 127 annotations with valid properties, values, or completely identical content. The open-source models notably struggle more compared to the proprietary \acp{llm}. While both open-source Llama models produce multiple valid outputs for properties and values with few-shot prompting, they fail to generate any valid predictions in the zero-shot setting. Therefore, these model runs are not displayed in the chart. Remarkably, GPT-4o outperforms GPT-3.5 by almost double, even in the zero-shot experiment, surpassing all other models by a great margin. In the few-shot cases, every third prediction from GPT-4o is identical to our annotated groundings, totaling 42 out of 127 instances. In some cases, The GPT-4o model even succeeded in precisely matching the annotated JSON-LD in a given conversation across a number of subsequent turns.

Table~\ref{tab:issues-frequency} provides a detailed analysis of the most common prediction issues and their relative frequencies for each model-prompt experiment. Examples for each issue type are listed in Table~\ref{tab:issue-types} in Appendix~\ref{sec:appendix-a}. Open-source models generally produce more invalid JSON-LD arrays and hallucinate properties and values that are not part of the system knowledge. All tested \acp{llm} tend to overpredict properties and values in zero-shot settings, even though these are grounded later in the conversation. Few-shot prompting can reduce excess properties and values, as well as counteract property deficits. However, in few-shot prompting, open-source models, particularly Llama-3-8B, tend to increase value deficits, becoming too hesitant to identify knowledge. This often results in empty JSON-LD arrays with generated statements such as ``The conversation does not mention any specific knowledge items from the system knowledge.''

Our findings corroborate existing benchmarks, highlighting the sophisticated reasoning abilities of state-of-the-art proprietary \acp{llm} such as GPT-4o in highly complex tasks. A similar task complexity-based \ac{llm} performance gap is also observable in the direct comparison of the MMLU and HumanEval benchmark scores between GPT-4o and Llama-3 \cite{hendrycks2020measuring,chen2021evaluating,openai2024gpt4o}. While Llama-70B performs competitively in the language-focused grounding act classification task, the superiority of GPT-4o becomes apparent in identifying knowledge when handling structured JSON-LD data and fragmented information from dialogue utterances. 

In short, when designing dialogue systems augmented with \acp{llm} to handle conversational grounding, smaller open-source models like Llama-3-8B, especially fine-tuned versions, seem to be generally sufficient for basic \ac{nlp} tasks such as detecting and classifying grounding-related dialogue acts. However, more complex tasks, such as identifying and integrating grounded knowledge from dialogue utterances with structured knowledge representations, require the use of more advanced and larger models like GPT-4o, which possess superior reasoning capabilities and proficiency in processing structured data formats.

\section{Conclusion and Future Work}
\label{sec:conclusion}
Our study examined \acp{llm} for handling grounding-related knowledge in information-sharing dialogues. We found that classifying grounding acts was feasible for both open- and closed-source \acp{llm}, with open-source \acp{llm} performing on par compared with leading proprietary ones. However, identifying grounded knowledge proved to be a distinctly more complex task. For the latter, the proprietary \acp{llm} had a competitive edge, and the open-source models underperformed due to their higher predisposition to generate erroneous output. The experiment results from our newly created dataset highlight common prediction issues and demonstrate how few-shot prompting can enhance model outputs, offering valuable insights to advance research on conversational grounding.

Future work should concentrate on developing \ac{llm}-based dialogue systems that handle conversational grounding through a multi-component pipeline approach for recognizing grounding-specific dialogue acts as well as grounded knowledge \cite{jokinen2024towards}. In previous studies, we have shown that \acp{llm} can augment dialogue systems by performing semantic parsing for conversational question answering over knowledge graphs \cite{schneider2024evaluating} and by verbalizing retrieved semantic triples into text responses \cite{schneider-etal-2024-comparative}. We believe conversational grounding is essential as it links the processes of semantic parsing of dialogue utterances, knowledge identification, and response generation, aligning the user's prior knowledge with the system’s available knowledge base while maintaining the relevance and coherence of conversations.

\section{Limitations}
Our study has certain limitations that should be acknowledged. First, the experiments are based on a relatively small dataset, consisting of only 26 information-seeking conversations and 669 dialogue turns collected in a controlled laboratory setting. While these conversations span five distinct domains, the findings should be interpreted with caution, as they may not generalize to larger or more diverse dialogue corpora.

Additionally, the grounded knowledge annotations in our study are represented using the JSON-LD syntax. We chose the JSON-LD format because it is widely used, and many \acp{llm} are trained to process JSON sequences effectively. However, it is important to recognize that other encoding formats, such as Turtle, RDF/XML, and N-Triples, may produce different performance results. Further, our experiments were restricted to the open-source Llama \cite{meta2024llama} and closed-source GPT \cite{openai2022chat,openai2024gpt4o} model families. It is advisable for future work to explore an even bigger variety of \acp{llm}, particularly those that are specifically trained on code and structured data like Codestral or Code Llama.

Lastly, conversational grounding in dialogue systems entails both the classification of grounding acts and the identification of grounded knowledge. While we have introduced and evaluated these tasks separately, incorporating our approach into an end-to-end evaluation could offer a more holistic understanding of end-to-end performance in more realistic dialogue scenarios.

\section{Ethical Considerations}
In our experiments, we used a publicly available dialogue dataset from \citet{schneider2023investigating} while ensuring that no personal identifying information of the participants was processed or disclosed. The information-seeking conversations from the dataset discuss only domain-specific knowledge from publicly accessible websites, such as Wikipedia. Moreover, to ensure optimal computing efficiency, evaluations of the Llama and GPT models were conducted on cloud computing platforms, with each inference run taking less than an hour.

\section*{Acknowledgements}
We would like to thank the anonymous reviewers for their helpful suggestions. Kristiina Jokinen acknowledges the support of Project JPNP20006 commissioned by the New Energy and Industrial Technology Development Organization (NEDO), Japan.

\bibliography{custom}

\newpage
\onecolumn
\appendix
\section{Appendix}
\label{sec:appendix-a}
\begin{figure*}[h]
\centering
  \includegraphics[width=0.85\textwidth]{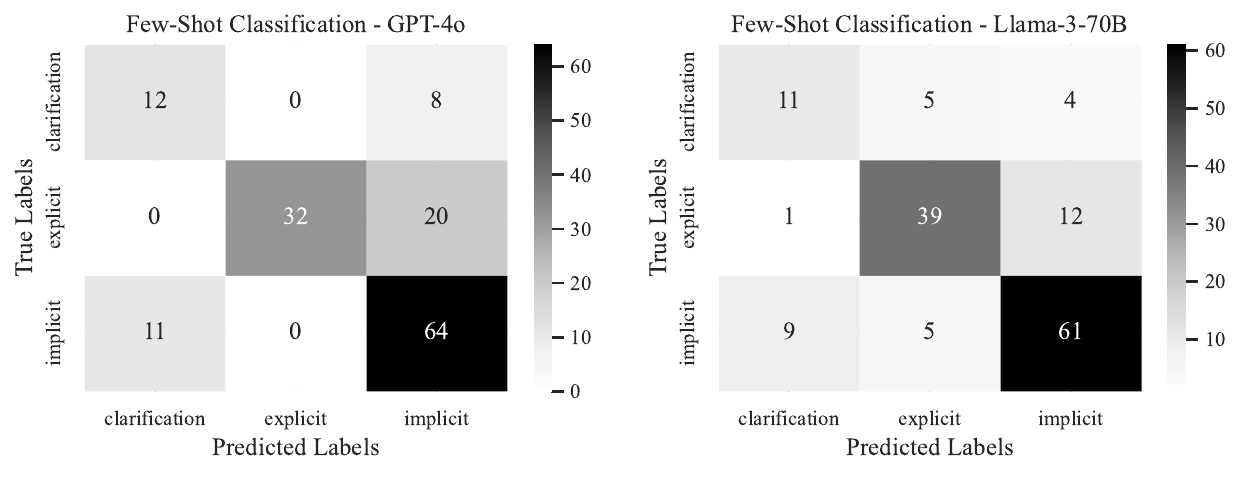}
  \caption{Confusion matrices for few-shot classification results of GPT-4o with three input utterances and Llama-3-70B with all input utterances.}
  \label{fig:heatmap-plot}
\end{figure*}

The Appendix provides one annotated conversation example (Table \ref{tab:dataset-example}), the model prompts in full length (Tables~\ref{tab:prompts-cls} and \ref{tab:prompts-gk}), an overview of common issue types identified in the predictions (Table~\ref{tab:issue-types}), and two confusion matrices of the classification results of the two best-performing model inference runs (Figure~\ref{fig:heatmap-plot}). 

\begin{table*}[h!]
\small
\centering
\begin{tabular}{p{0.375\textwidth}p{0.055\textwidth}p{0.475\textwidth}}
\hline
\textbf{Dialogue Utterances} & \textbf{Dialogue Act} & \textbf{Grounded Knowledge} \\
\hline
S: What is your dataset about? & \centering - & - \\
P: it contains information about 11341 historical figures, including their full name, sex, birth year, city, country, continent, occupation, historical popularity index (HPI). The HPI represents the degree of this person's online popularity & \centering - & - \\
S: Who is the most popular? & \centering implicit & [\{"@context": ["http://www.w3.org/ns/csvw", \{"schema": "http://schema.org"\}], "@id": "http://example.org/historical-figures", "url": "historical-figures.csv", \textcolor{blue}{"schema:description": "The table contains information about historical figures", "tableSchema": \{"columns": [\{"name": "full\_name", "datatype": "string"\}, \{"name": "sex", "datatype": "string"\}, \{"name": "birth\_year", "datatype": "integer"\}, \{"name": "city", "datatype": "string"\}, \{"name": "country", "datatype": "string"\}, \{"name": "continent", "datatype": "string"\}, \{"name": "occupation", "datatype": "string"\}, \{"name": "historical\_popularity\_index", "datatype": "float"\}], "primaryKey": "full\_name"\}\}}] \\
P: Aristotle, who is from Greece and has a largest HPI value: 31.9938. & \centering - & - \\
S: I see, is there Socrate in the dataset? & \centering explicit & [\{"@context": ["http://www.w3.org/ns/csvw", \{"schema": "http://schema.org"\}], "@id": "http://example.org/historical-figures", "url": "historical-figures.csv", "schema:description": "The table contains information about historical figures", "tableSchema": \{"columns": [\{"name": "full\_name", "datatype": "string"\}, \{"name": "sex", "datatype": "string"\}, \{"name": "birth\_year", "datatype": "integer"\}, \{"name": "city", "datatype": "string"\}, \{"name": "country", "datatype": "string"\}, \{"name": "continent", "datatype": "string"\}, \{"name": "occupation", "datatype": "string"\}, \{"name": "historical\_popularity\_index", "datatype": "float", \textcolor{blue}{"maximum": 31.9938}\}], "primaryKey": "full\_name"\}\}, \textcolor{blue}{\{"@type": "schema:Person", "full\_name": "Aristotle", "country": "Greece", "historical\_popularity\_index": 31.9938\}}] \\
P: Yes, Socrate is in the dataset. & \centering - & - \\
S: What is is popularity index? & \centering implicit & [\{"@context": ["http://www.w3.org/ns/csvw", \{"schema": "http://schema.org"\}], "@id": "http://example.org/historical-figures", "url": "historical-figures.csv", "schema:description": "The table contains information about historical figures", "tableSchema": \{"columns": [\{"name": "full\_name", "datatype": "string"\}, \{"name": "sex", "datatype": "string"\}, \{"name": "birth\_year", "datatype": "integer"\}, \{"name": "city", "datatype": "string"\}, \{"name": "country", "datatype": "string"\}, \{"name": "continent", "datatype": "string"\}, \{"name": "occupation", "datatype": "string"\}, \{"name": "historical\_popularity\_index", "datatype": "float", "maximum": 31.9938\}], "primaryKey": "full\_name"\}\}, \{"@type": "schema:Person", "full\_name": "Aristotle", "country": "Greece", "historical\_popularity\_index": 31.9938\}, \textcolor{blue}{\{"@type": "schema:Person", "full\_name": "Socrates"\}}] \\
P: Historical popularity index (HPI) is metric that aggregates information on a biography’s online popularity. It aggregates information on the age and attention received by biographies in multiple language editions of Wikipedia to provide a summary statistic of their global popularity. & \centering -  & - \\
\hline 
\end{tabular}
\caption{Example of dialogue excerpt from the history domain with annotated grounding dialogue acts and grounded knowledge in JSON-LD format. Seeker (S) and provider (P) roles are abbreviated for each turn. Utterances are taken from the dialogue logs and may contain spelling errors. Newly grounded knowledge is displayed in \textcolor{blue}{blue} color.}
\label{tab:dataset-example}
\end{table*}

\begin{table*}[h]
\small
\centering
\begin{tabular}{p{0.95\textwidth}}
\hline
\textbf{Grounding Act Classification Prompt}\\
\hline
\textbf{Zero-Shot}\\
\hline
\verb|SYSTEM:| Predict the grounding label for the last response in the 'Input Dialogue:'. The label indicates whether the knowledge in the dialogue was accepted. Choose one of the following labels:
\newline explicit: The response confirms understanding or acceptance (e.g., 'okay', 'thanks', 'alright', 'nice') without seeking clarification.
\newline clarification: The response seeks clarification about a previous dialogue snippet.
\newline implicit: The response moves the conversation forward without explicitly confirming or seeking clarification.
\\
\hline
\textbf{Few-Shot}\\
\hline
\verb|SYSTEM:| Predict the grounding label for the last response in the 'Input Dialogue:'. The label indicates whether the knowledge in the dialogue was accepted. Choose one of the following labels:
\newline explicit: The response confirms understanding or acceptance (e.g., 'okay', 'thanks', 'alright', 'nice') without seeking clarification.
\newline clarification: The response seeks clarification about a previous dialogue snippet.
\newline implicit: The response moves the conversation forward without explicitly confirming or seeking clarification.
\newline \newline
\verb|USER:| Input Dialogue: \newline seeker: Can you give me some information about your dataset? \newline provider: My dataset includes information on buildings of Gothic architecture. \newline seeker: How tall is the Cologne Cathedral?
\newline
\verb|ASSISTANT:| Output Label: implicit
\newline \newline
\verb|USER:| Input Dialogue: \newline provider: Monitors have different attributes like size or panel technology. \newline provider: There are some with an aspect ratio of 21:9. \newline seeker: What is aspect ratio?
\newline
\verb|ASSISTANT:| Output Label: clarification
\newline \newline
\verb|USER:| Input Dialogue: \newline provider: An elephant's average lifespan is around 65 years. \newline seeker: I see, good to know.
\newline
\verb|ASSISTANT:| Output Label: explicit\\
\hline
\end{tabular}
\caption{Overview of applied zero-shot and few-shot prompts for classification.}
\label{tab:prompts-cls}
\end{table*}

\begin{table*}[h]
\small
\centering
\begin{tabular}{p{0.95\textwidth}}
\hline
\textbf{Grounded Knowledge Identification Prompt}\\
\hline
\textbf{Zero-Shot}\\
\hline
\verb|SYSTEM:| Your task is to identify the knowledge items that have been grounded by the conversation partners in the 'Input~Dialogue'. The items of mutually grounded knowledge must be explicitly mentioned in the dialogue. Based on the complete set of 'System Knowledge', your task is to generate the subset of knowledge items that have been grounded so far. Ensure that the output is a valid JSON-LD structure (an array of JSON objects) and only include knowledge items from the formatted 'System Knowledge'.
\newline \\
\hline
\textbf{Few-Shot}\\
\hline
\verb|SYSTEM:| Your task is to identify the knowledge items that have been grounded by the conversation partners in the 'Input~Dialogue'. The items of mutually grounded knowledge must be explicitly mentioned in the dialogue. Based on the complete set of 'System Knowledge', your task is to generate the subset of knowledge items that have been grounded so far. Ensure that the output is a valid JSON-LD structure (an array of JSON objects) and only include knowledge items from the formatted 'System Knowledge'.
\newline \newline
\verb|USER:| System Knowledge: [\{"@context": ["http://www.w3.org/ns/csvw", \{"schema": "http://schema.org"\}], "@id": "http://example.org/american-presidents", "url": "american-presidents.csv", "schema:description": "The table contains information about American presidents", "tableSchema": \{"columns": [\{"name": "name", "datatype": "string"\}, \{"name": "term", "datatype": "string"\}, \{"name": "party", "datatype": "string"\}, \{"name": "election\_year", "datatype": "integer"\}]\}, "primaryKey": "name"\}, \{"@type": "schema:Person", "name": "Barack Obama", "party": "Democratic"\}]
\newline Input Dialogue:
\newline seeker: Can you give me an example entry from your dataset?
\newline provider: One of the presidents in the list is Barack Obama.
\newline seeker: Thanks. What party does he belong to?
\newline \newline
\verb|ASSISTANT:| Output JSON-LD: [\{"@context": ["http://www.w3.org/ns/csvw", \{"schema": "http://schema.org"\}], "@id": "http://example.org/american-presidents", "url": "american-presidents.csv", "schema:description": "The table contains information about American presidents", "tableSchema": \{"columns": [\{"name": "name", "datatype": "string"\}]\}, "primaryKey": "name"\}, \{"@type": "schema:Person", "name": "Barack Obama"\}]
\newline \newline
\verb|USER:| System Knowledge: [\{"@context": ["http://www.w3.org/ns/csvw", \{"schema": "http://schema.org"\}], "@id": "http://example.org/greek-islands", "url": "greek-islands.csv", "schema:description": "The table contains information about islands in Greece", "tableSchema": \{"columns": [\{"name": "island", "datatype": "string"\}, \{"name": "area\_in\_km2", "datatype": "integer", "minimum": 64, "maximum": 8336\}, \{"name": "cluster", "datatype": "string"\}]\}, "primaryKey": "island"\}, \{"@type": "schema:Place", "island": "Crete", "area\_in\_km2": 8336, "cluster": "Cretan"\}, \{"@type": "schema:Place", "island": "Alonnisos", "area\_in\_km2": 64, "cluster": "Sporades"\}, \{"@type": "schema:Place", "island": "Lesbos", "area\_in\_km2": 1633, "cluster": "North Aegean Islands"\}]
\newline Input Dialogue:
\newline provider: My dataset contains information on Greek islands. For example, there is Crete with an area of 8336 square kilometers.
\newline provider: That makes it the largest island in Greece.
\newline seeker: Which one is the smallest and what is its area?
\newline \newline
\verb|ASSISTANT:| Output JSON-LD: [\{"@context": ["http://www.w3.org/ns/csvw", \{"schema": "http://schema.org"\}], "@id": "http://example.org/greek-islands", "url": "greek-islands.csv", "schema:description": "The table contains information about islands in Greece", "tableSchema": \{"columns": [\{"name": "island", "datatype": "string"\}, \{"name": "area\_in\_km2", "datatype": "integer", "maximum": 8336\}]\}, "primaryKey": "island"\}, \{"@type": "schema:Place", "island": "Crete", "area\_in\_km2": 8336\}]
\newline \newline
\verb|USER:| System Knowledge: [\{"@context": ["http://www.w3.org/ns/csvw", \{"schema": "http://schema.org"\}], "@id": "http://example.org/android-smartphones", "url": "android-smartphones.csv", "schema:description": "The table contains information about Android smartphones", "tableSchema": \{"columns": [\{"name": "model", "datatype": "string"\}, \{"name": "developer", "datatype": "string"\}, \{"name": "release\_year", "datatype": "integer", "minimum": 2008, "maximum": 2024\}, \{"name": "android\_version", "datatype": "string"\}]\}, "primaryKey": "model"\}, \{"@type": "schema:Product", "model": "HTC Dream", "developer:": "HTC", "release\_year": "2008"\}, \{"@type": "schema:Product", "model": "LG Wing", "developer:": "LG", "release\_year": "2020", "android\_version": "Android 10"\}, \{"@type": "schema:Product", "release\_year": "2024"\}]
\newline Input Dialogue:
\newline provider: I can provide technical information about Android smartphones.
\newline provider: One column contains data about the model and another specifies its release year.
\newline seeker: I see, good to know.
\newline \newline
\verb|ASSISTANT:| Output JSON-LD: [\{"@context": ["http://www.w3.org/ns/csvw", \{"schema": "http://schema.org"\}], "@id": "http://example.org/android-smartphones", "url": "android-smartphones.csv", "schema:description": "The table contains information about Android smartphones", "tableSchema": \{"columns": [\{"name": "model", "datatype": "string"\}, \{"name": "release\_year", "datatype": "integer"\}]\}, "primaryKey": "model"\}]
\newline
\\
\hline
\end{tabular}
\caption{Overview of applied zero-shot and few-shot prompts for grounding knowledge.}
\label{tab:prompts-gk}
\end{table*}

\begin{table*}[h]
\small
\centering
\begin{tabular}{p{1.4cm}p{3.3cm}p{9.7cm}}
\hline
\textbf{Issue Type} & \textbf{Definition} & \textbf{Example}\\
\hline
Invalid \newline JSON-LD & Prediction is overall not a valid JSON-LD structure. & \verb|PRED:| [\{"@type": "schema:CreativeWork", "category": ["films", "series", "literature"]\textcolor{red}{\}]} \break \textcolor{red}{$\rightarrow$ JSON syntax elements are missing.} \\
\hline
Property \newline Hallucination & Predicted property is not part of system knowledge. & \verb|PRED:|  \{"@type": "schema:Place", "name": "English Garden", "state": "Bavaria", \textcolor{red}{"city"}: "Munich", "summary": "The English Garden is a large public park in Munich, Bavaria. It is one of the world's largest urban public parks, larger than New York's Central Park. The park is known for its beer gardens, streams, lakes, and Japanese teahouse."\} \break \textcolor{red}{$\rightarrow$ English Garden was mentioned in the dialogue but is not part of the system knowledge. The "city" property does not exist.} \\
\hline 
Value \newline Hallucination & Predicted value is not part of system knowledge. & \verb|PRED:| [\{"@context": ["http://www.w3.org/ns/csvw", \{"schema": "http://schema.org"\}], "@id": "http://example.org/time-travel-media", "url": "time-travel-media.csv", "schema:description": "The table contains information about time travel media", "tableSchema": \{"columns": [\{"name": "year", "datatype": "string"\}, \{"name": "title", "datatype": "string"\}, \{"name": \textcolor{red}{"author"}, "datatype": "string"\}, \{"name": \textcolor{red}{"short text description"}, "datatype": "string"\}, \{"name": "category", "datatype": "string"\}]\}, "primaryKey": ["year", "title", "category"]\}] \break \textcolor{red}{$\rightarrow$ System knowledge contains the columns "creators" and "description" but "author" and "short text description" are hallucinated.} \\
\hline 
Property \newline Excess & Predicted property is part of system knowledge but grounded too early. & \verb|PRED:| [\{"@context": ["http://www.w3.org/ns/csvw", \{"schema": "http://schema.org"\}], "@id": "http://example.org/football-records", "url": "football-records.csv", "schema:description": "The table contains information about football records", "tableSchema": \{"columns": [\{\textcolor{red}{"name"}: "record", \textcolor{red}{"datatype"}: "string"\}]\}, "primaryKey": "record"\}] \break \textcolor{red}{$\rightarrow$ Properties should have been grounded at a later stage.} \\
\hline  
Property \newline Deficit & Property from system knowledge was not identified. & \verb|PRED:| [\{"@context": ["http://www.w3.org/ns/csvw", \{"schema": "http://schema.org"\}], "@id": "http://example.org/nature-parks", "url": "nature-parks.csv", "schema:description": "The table contains information about nature parks in Germany", "tableSchema": \{"columns": [\{"name": "name", "datatype": "string"\}, \{\textcolor{red}{"name"}: "state", \textcolor{red}{"datatype"}: "string"\}, \{\textcolor{red}{"name"}: "year", \textcolor{red}{"datatype"}: "integer"\}, \{\textcolor{red}{"name"}: "area\_in\_km2", \textcolor{red}{"datatype"}: "integer"\}, \{\textcolor{red}{"name"}: "summary", \textcolor{red}{"datatype"}: "string"\}], "primaryKey": "name"\}\}, \{"@type": "schema:Place", "name": "Altmühl Valley Nature Park"\}] \break \textcolor{red}{$\rightarrow$ Highlighted properties were not grounded.} \\
\hline
Value \newline Excess & Predicted value is part of system knowledge but grounded too early. & \verb|PRED:| [\{"@context": ["http://www.w3.org/ns/csvw", \{"schema": "http://schema.org"\}], "@id": "http://example.org/nature-parks", "url": "nature-parks.csv", "schema:description": "The table contains information about nature parks in Germany", "tableSchema": \{"columns": [\{"name": \textcolor{red}{"name"}, "datatype": \textcolor{red}{"string"}\}, \{"name": \textcolor{red}{"state"}, "datatype": \textcolor{red}{"string"}\}, \{"name": \textcolor{red}{"year"}, "datatype": \textcolor{red}{"integer"}\}, \{"name": \textcolor{red}{"area\_in\_km2"}, "datatype": \textcolor{red}{"integer"}\}, \{"name": \textcolor{red}{"summary"}, "datatype": \textcolor{red}{"string"}\}], "primaryKey": "name"\}\}] \break \textcolor{red}{$\rightarrow$ Values should have been grounded at a later stage.} \\
\hline 
Value \newline Deficit & Value from system knowledge was not identified. & \verb|PRED:| [\{"@context": ["http://www.w3.org/ns/csvw", \{"schema": "http://schema.org"\}], "@id": "http://example.org/historical-figures", "url": "historical-figures.csv", "schema:description": "The table contains information about historical figures", "tableSchema": \{"columns": [\{"name": "full\_name", "datatype": "string"\}, \{"name": "birth\_year", "datatype": "integer", "minimum": \textcolor{red}{-3500}, "maximum": \textcolor{red}{2005}\}], "primaryKey": "full\_name"\}\}, \{"@type": \textcolor{red}{"schema:Person"}, "birth\_year": \textcolor{red}{-3500}\}, \{"@type": \textcolor{red}{"schema:Person"}, "birth\_year": \textcolor{red}{2005}\}] \break \textcolor{red}{$\rightarrow$ Highlighted values were not grounded.} \\
\hline
\end{tabular}
\caption{Overview of six identified issue types with examples from generated model predictions (PRED). The manifestation of issues are highlighted in \textcolor{red}{red} color.}
\label{tab:issue-types}
\end{table*}

\end{document}